\documentclass[graybox]{SNmult}
\usepackage[backend=biber]{biblatex}
\usepackage{type1cm}        
\usepackage{makeidx}         
\usepackage{graphicx}        
\usepackage{multicol}        
\usepackage[bottom]{footmisc}

\usepackage{newtxtext}       %
\usepackage[varvw]{newtxmath}       
\usepackage{tabularx}
\usepackage{array}
\usepackage{booktabs}
\usepackage{hyperref}
\makeindex             
\begin{document}

\title*{ AI-Enabled Serious Games: Integrating Intelligence and Adaptivity in Training Systems}
\author{Priyamvada Tripathi\orcidID{0009-0005-5070-7420} and\\ Bill Kapralos\orcidID{0000-0003-0434-3847}}
\institute{Priyamvada Tripathi \at Durham College, 2000 Simcoe Street North
Oshawa, Ontario, Canada. L1G 0C5, \email{priyamvada.tripathi@durhamcollege.ca}
\and Bill Kapralos \at Ontario Tech University, 2000 Simcoe Street North, Oshawa, Ontario, Canada. L1G 0C5 \email{bill.kapralos@ontariotechu.ca}}
%
%
\maketitle
\abstract*{Serious games are widely used for learning and training across domains such as healthcare, defense, and education. Persistent challenges remain, however, including static scenario design, authoring bottlenecks, limited learner modeling, and difficulty implementing meaningful real-time instructional adaptation. Recent advances in artificial intelligence (AI) introduce novel capabilities such as dynamic scenario variation, contextual feedback, adaptive pacing, and learner-state modeling that may help address some of these limitations. At the same time, integrating AI into serious games raises important questions related to validity, transparency, system control, and learner trust. This chapter examines how contemporary AI approaches may support real-time instructional adaptation in serious games. It distinguishes between instructional intelligence, which is defined as a system’s capacity to infer learner knowledge and reason about pedagogically appropriate responses, and adaptivity, which is defined as the ability to modify instructional actions during interaction. A historical synthesis of adaptive learning systems is presented, tracing developments from early computer-assisted instruction through intelligent tutoring systems (ITS), dynamic difficulty adjustment (DDA), authoring platforms, learning analytics, and recent AI-enabled architectures. Building on this perspective, the chapter discusses how large language models (LLMs), reinforcement learning (RL), and agent-based architectures may contribute to more integrated forms of intelligence and adaptivity in serious games. It also highlights practical and research challenges associated with AI-enabled systems, including explainability, validation, computational cost, and the limited empirical evidence regarding long-term learning outcomes in AI-enabled serious games.
}

\abstract{Serious games are widely used for learning and training across domains such as healthcare, defense, and education. Persistent challenges remain, however, including static scenario design, authoring bottlenecks, limited learner modeling, and difficulty implementing meaningful real-time instructional adaptation. Recent advances in artificial intelligence (AI) introduce novel capabilities such as dynamic scenario variation, contextual feedback, adaptive pacing, and learner-state modeling that may help address some of these limitations. At the same time, integrating AI into serious games raises important questions related to validity, transparency, system control, and learner trust. This chapter examines how contemporary AI approaches may support real-time instructional adaptation in serious games. It distinguishes between instructional intelligence, defined as a system’s capacity to infer learner knowledge and reason about pedagogically appropriate responses, and adaptivity, defined as the ability to modify instructional actions during interaction. A historical synthesis of adaptive learning systems is presented, tracing developments from early computer-assisted instruction through intelligent tutoring systems (ITS), dynamic difficulty adjustment (DDA), authoring platforms, learning analytics, and recent AI-enabled architectures. Building on this perspective, the chapter discusses how large language models (LLMs), reinforcement learning (RL), and agent-based architectures may contribute to more integrated forms of intelligence and adaptivity in serious games. It also highlights practical and research challenges associated with AI-enabled systems, including explainability, validation, computational cost, and the limited empirical evidence regarding long-term learning outcomes in AI-enabled serious games.\newline\indent
}

\section{Introduction}
\label{sec:1}
Serious games are interactive digital environments designed for learning, training, and behavior change \cite{ritterfeld_serious_2009}. They have demonstrated learning benefits across multiple domains, including healthcare \cite{graafland_systematic_2012}\cite{kapralos_levels_2017}, defense \cite{smith_long_2010}\cite{zyda_visual_2005}, and education \cite{plass_foundations_2015}\cite{radianti_systematic_2020}. According to Bloom’s Taxonomy \cite{anderson_blooms_1994}, learning encompasses three domains: cognitive (acquiring and applying knowledge), affective (developing attitudes, values, and motivation), and psychomotor (building physical and procedural skills). Serious games have predominantly focused on the cognitive and affective domains, with more limited application to psychomotor skills. This gap reflects both instructional design constraints and the difficulty of accurately simulating haptic feedback (sense of touch and physical resistance) which requires expensive and complex hardware \cite{kapralos_immersive_2024}. Persistent technical challenges further constrain serious game effectiveness, including authoring constraints that restrict adaptivity to predefined scenarios \cite{bellotti_assessment_2013}\cite{lopes_adaptivity_2011}, difficulty accurately modeling learner knowledge states \cite{desmarais_review_2012}, challenges implementing real-time instructional adaptation \cite{fu_examining_2025}\cite{sottilare_design_2013}, and inconsistent evidence of skill transfer beyond the training environment \cite{van_oostendorp_editorial_2022}.

Recent advances in applied artificial intelligence (AI) offer mechanisms that may address some of these longstanding limitations. Learner modeling refers to the computational process of tracking and updating a system's representation of what an individual learner knows, misunderstands, and still needs to learn. Prior work in adaptive instructional systems and intelligent tutoring has demonstrated how AI-driven learner modeling and adaptive control can support individualized feedback, pacing, and progression without relying on rigid pre-authored rules \cite{brusilovsky_adaptive_2001}\cite{shute_adaptive_2012}\cite{sottilare_design_2013}\cite{woolf_building_2008}. Emerging AI-enabled approaches have begun to support dynamic difficulty adjustment (DDA), adaptive feedback, and real-time learner state inference grounded in interaction data rather than explicitly authored representations \cite{ferguson_ai-induced_2022}\cite{lopes_adaptivity_2011}. Broader reviews of AI in education confirm that adaptive and personalized learning systems are among the most widely studied applications across educational contexts \cite{fu_examining_2025}\cite{wouters_meta-analysis_2013}. Collectively, this research suggests that AI may extend serious games toward more personalized, adaptive learning trajectories \cite{luckin_intelligence_2016} including through the design of systems that integrate human and machine intelligence to support student learning \cite{holstein_student_2018}.

However, the integration of AI into serious games warrants caution. Historically, emerging technologies have been positioned as solutions to persistent limitations in learning and training systems, often without consistent empirical support. This pattern was evident in the adoption of virtual reality (VR) and immersive technologies, where increases in technical sophistication, particularly visual fidelity and immersion, were often assumed to produce corresponding gains in effectiveness. Empirical evidence indicates that such increases alone do not reliably translate into improved learning or training outcomes \cite{hamstra_reconsidering_2014}\cite{kapralos_levels_2017}. Systematic reviews and meta-analyses of AI in education have similarly shown mixed learning gains \cite{fu_examining_2025}\cite{holmes_artificial_2023}\cite{tlili_investigating_2025}\cite{yeo_effects_2025}\cite{zawacki-richter_systematic_2019}. Accordingly, claims regarding the effectiveness of AI in serious games warrant a measured, evidence-based perspective. Without clear empirical and pedagogical grounding, AI-enabled serious games risk creating the appearance of personalization or intelligent instruction while failing to support meaningful learning or skill transfer, similar to the pattern observed in earlier waves of computer-assisted instruction and educational technology adoption \cite{cuban_oversold_2001}\cite{means_effectiveness_2013}.

This chapter treats instructional intelligence and adaptivity as two distinct system properties. Instructional intelligence refers to a system's capacity to infer learner state and reason about pedagogically appropriate responses, encompassing computational models of what the learner knows, what is misunderstood, and what instructional interventions will address specific deficits \cite{brusilovsky_adaptive_2001}\cite{woolf_building_2008}. Adaptivity refers to the system's ability to modify instructional actions at runtime. That is, adaptivity is the ability to alter system behavior during an active learning session, without pausing or rebuilding the system in response to learner interaction \cite{brusilovsky_adaptive_2001}\cite{lopes_adaptivity_2011}\cite{shute_adaptive_2012}. Addressing these dimensions requires articulating the architectural considerations and implementation challenges involved in integrating AI within established learning and training evaluation frameworks for serious games.

The remainder of this chapter is organized as follows. Section 2 introduces the distinction between instructional intelligence and adaptivity, treats them as independent system properties, and traces their historical evolution across six phases of adaptive training systems spanning 1969 to the present. Section 3 analyzes how large language models (LLMs), reinforcement learning (RL), and agent-based architectures can support more integrated forms of intelligence and adaptivity in serious games. Section 4 synthesizes these perspectives to identify key design considerations, implementation challenges, and research directions for responsible AI integration in serious games and immersive simulation-based training.

\section{Background: The Evolution Toward Intelligent-Adaptive Integration}
\label{sec:2}
While serious games have made significant advances in instructional design, validation methodology, and immersive technology, persistent challenges remain. Static scenario designs such as linear progressions, branching decision trees, or discrete difficulty tiers create fundamental limitations. First, learners may exhaust scenarios after only a few repetitions, which can reduce learning gains over time \cite{kapralos_immersive_2024}\cite{torres_gamification_2025}. Second, scenarios are typically encoded at design time with the involvement of content experts and game developers, requiring significant effort and consequently limiting the frequency with which content can be updated \cite{bellotti_assessment_2013}\cite{lopes_adaptivity_2011}. Although some preliminary work has investigated authoring platforms that allow educators to create or modify scenarios directly \cite{torres_gamification_2025}, these remain highly constrained and are not yet widely deployed. Third, generic or canned feedback fails to address learner-specific misconceptions \cite{hattie_power_2007}\cite{vanlehn_behavior_2006}. Fourth, authoring bottlenecks require many hours per scenario, limiting production systems to small content libraries \cite{bellotti_assessment_2013}\cite{torres_gamification_2025}. Fifth, psychomotor skill development remains difficult to support in virtual environments, given that accurately simulating the sense of touch requires expensive and complex haptic equipment \cite{kapralos_immersive_2024}\cite{kapralos_levels_2017}. Sixth, one-size-fits-all experiences that fail to account for diverse learner backgrounds, abilities, and prior knowledge constitute an under-examined limitation of current systems \cite{kapralos_immersive_2024}.

In the face of these limitations and equipped with advances in immersive technology and computing power, some researchers pursued increased perceptual fidelity and realism in serious games and virtual simulations. The assumption was that more realistic environments would improve training effectiveness of serious games and overcome some of these limitations. However, empirical evidence contradicted this assumption and has shown that increased fidelity (sometimes called the fidelity fallacy) fails to reliably improve learning outcomes \cite{kapralos_levels_2017}. In some cases, higher levels of fidelity increase extraneous cognitive load (the mental effort spent processing the environment rather than the learning content) without proportional learning benefit while escalating development costs \cite{kapralos_levels_2017}. These studies collectively suggest that the bottlenecks constraining serious game effectiveness cannot be addressed simply by improving content quantity or visual fidelity.

Research has therefore addressed these challenges by integrating instructional intelligence and adaptivity which are two conceptually distinct system properties that have proven difficult to realize simultaneously within unified architectures across more than five decades of research \cite{brusilovsky_adaptive_2001}\cite{sottilare_design_2013}\cite{vanlehn_relative_2011}. Instructional intelligence requires detailed models of what a learner understands, where they are making errors, and how their knowledge is progressing including multidimensional models of conceptual knowledge, procedural skills, and learning progress built into the training systems \cite{desmarais_review_2012}\cite{koedinger_instructional_2013}\cite{vanlehn_behavior_2006}. At the very least, intelligent serious games should be capable of diagnosing the underlying causes of observable performance, such as distinguishing misconceptions (incorrect beliefs about how something works) from procedural errors (mistakes in executing a known procedure), and systematic gaps from transient mistakes \cite{vanlehn_relative_2011}. Moreover, they should have the ability to reason pedagogically about appropriate instructional responses grounded in learning science \cite{chi_self-explanations_1989}\cite{koedinger_exploring_2007}\cite{vanlehn_behavior_2006}. This includes selecting feedback types aligned with error categories \cite{hattie_power_2007}\cite{shute_focus_2008}, and sequencing tasks to support scaffolding. Scaffolding is a teaching approach that provides structured support calibrated to the learner's current ability, progressively withdrawing that support as competence grows. It is aimed at alignment with the zone of proximal development, defined as the range of tasks a learner can accomplish with guidance but not yet independently \cite{vygotsky_mind_1978}\cite{wood_role_1976}. Ideally, instructional support in intelligent serious games should also be calibrated to the learner's inferred competence rather than to surface performance thresholds \cite{collins_cognitive_2018}\cite{pea_social_2004}.

By contrast, adaptivity denotes a system's ability to modify instructional actions dynamically at runtime and involves adjusting task difficulty, pacing, sequencing, scaffolding intensity, and feedback timing in response to learner interaction \cite{hunicke_ai_2004}\cite{lopes_adaptivity_2011}. Adaptivity operates on interaction-relevant timescales: seconds to minutes within scenarios for immediate difficulty adjustment, minutes to hours across scenarios for task sequencing, and hours to days for curriculum-level progression \cite{shute_adaptive_2012}. Crucially, adaptive behavior does not inherently require deep diagnostic understanding: systems may respond rapidly to simple performance signals such as accuracy rates, completion times, and error frequencies without modeling the cognitive structures that produced them \cite{sweetser_gameflow_2005}\cite{westera_artificial_2020}. Thus, adaptivity can be present without intelligence, that is a system can change what it presents to the learner without truly understanding why the learner is struggling.

\subsection{Instructional Intelligence and Adaptivity as Independent System Properties}
In the context of computer-based training systems, instructional intelligence and adaptivity are better understood as independent rather than sequential dimensions. This implies that a system can be strong on one dimension while still being weak on the other and improving one dimension does not automatically improve the other. Therefore, a system can exhibit (1) high instructional intelligence with low adaptivity (diagnosing learner states accurately but responding slowly or inflexibly), (2) high adaptivity with low intelligence (adjusting rapidly based on shallow performance signals without deeper diagnostic insight), (3) low capability on both dimensions (static content with fixed progression), or (4) high capability on both dimensions simultaneously, at least in principle. This fourth configuration (high intelligence-high adaptivity) has proven extraordinarily difficult to achieve in practice, producing persistent design trade-offs that have shaped the history of adaptive training systems \cite{brusilovsky_user_2007}\cite{sottilare_design_2013}\cite{vanlehn_relative_2011}. The architectural reason for this difficulty is not arbitrary: the computational methods required for deep learner modeling demanded significant processing time that was incompatible with real-time interaction, while systems designed for rapid response lacked computational resources for complex reasoning about learner cognition. Fig. 1  illustrates this two-dimensional space with representative systems from each phase of development described below.
\begin{figure}
    \centering
    \includegraphics[width=1\linewidth]{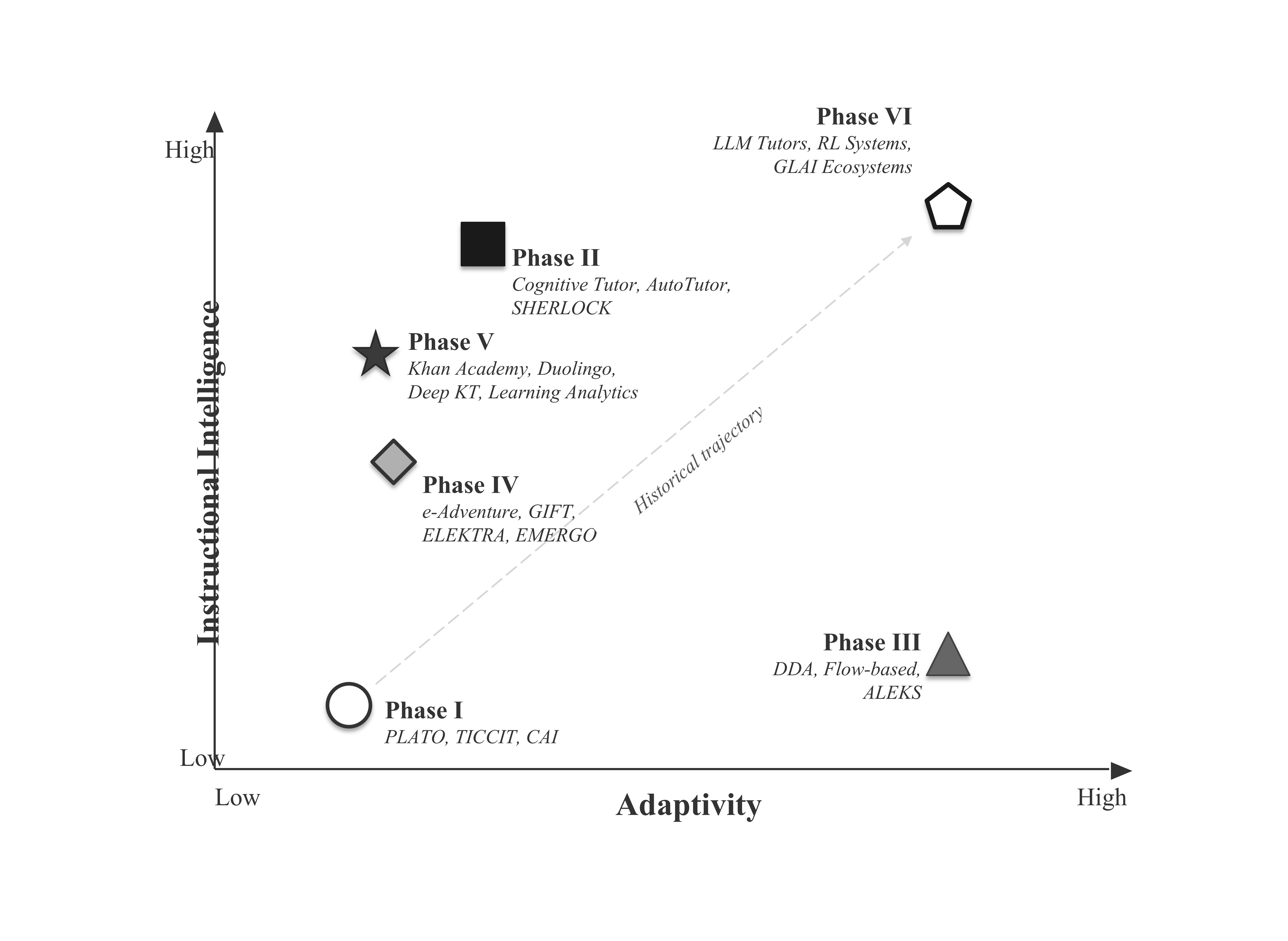}
    \caption{Intelligence–Adaptivity space for serious games and adaptive training systems. The vertical axis represents instructional intelligence (low to high); the horizontal axis represents adaptivity (low to high). Axis positions reflect relative capability rankings derived from the empirical literature cited in this chapter, not absolute measurements. Systems are positioned ordinally based on documented learner modeling depth, diagnostic specificity, adaptation timescale, and policy sophistication. Note. Individual system placements represent approximate ordinal rankings for illustrative purposes and should not be interpreted as precise quantitative measurements.}
     \Description{A conceptual two-dimensional plot with instructional intelligence on the vertical axis and adaptivity on the horizontal axis. Systems are positioned qualitatively across the space based on increasing intelligence and adaptivity, illustrating relative progression from simple rule-based systems to advanced adaptive, model-driven systems.}
    \label{fig:intelligence-adaptivity}
    \label{fig:placeholder}
\end{figure}

\subsection{A Historical Arc of Intelligence and Adaptivity in Serious Games}
To contextualize the significance of AI integration within serious games and virtual training, this section synthesizes a chronological progression documenting how computational learning systems have attempted to integrate instructional intelligence and enable runtime adaptation to individual learner performance, knowledge states, and instructional needs \cite{desmarais_review_2012}\cite{sottilare_design_2013}\cite{vanlehn_relative_2011}. This arc, traced through six phases spanning 1969 to the present, reveals that AI integration represents a qualitative shift in where intelligence resides within system architectures, with deep implications for capabilities, evaluation methodologies, deployment requirements, and research priorities (Table 1).

\begin{table*}[t]
\centering
\caption{Representative systems across six phases of intelligence and adaptivity arc in serious games}
\label{tab:historical_arc}

\scriptsize
\setlength{\tabcolsep}{3pt}
\renewcommand{\arraystretch}{1.15}

\begin{tabularx}{\textwidth}{>{\raggedright\arraybackslash}p{2.4cm}
>{\raggedright\arraybackslash}X
>{\raggedright\arraybackslash}X
>{\raggedright\arraybackslash}X
>{\raggedright\arraybackslash}p{2.6cm}}
\toprule
\textbf{Phase \& Representative Systems} &
\textbf{Instructional Intelligence Capabilities} &
\textbf{Adaptivity Capabilities} &
\textbf{Key Limitations}
 \\
\midrule

\textbf{Phase I (1969--1980s)}: PLATO, TICCIT, CAI drill-and-practice \cite{alessi_computer-based_1985}\cite{rubinoff_wide_1976}\cite{bunderson_work_1981}&
Minimal: simple numerical measures such as accuracy, response time, and error counts. No learner modeling or diagnostic reasoning. &
Reactive IF--THEN rules and threshold-based adjustments to speed or complexity. No anticipation or planning. &
Cannot diagnose causes of errors. No distinction between conceptual and procedural mistakes. Purely behaviorist. \\
\addlinespace

\textbf{Phase II (1980s--1990s)}: Cognitive Tutors, AutoTutor, SHERLOCK, SimStudent, Betty's Brain \cite{anderson_cognitive_1995}\cite{corbett_knowledge_1995}\cite{graesser_autotutor_2004}&
High: explicit learner models, production rules, misconception diagnosis, and cognitive task analysis. &
Low--medium: model tracing and probabilistic inference with limited real-time adaptation. &
Knowledge-engineering bottleneck, brittle models, and high computational cost. \\
\addlinespace

\textbf{Phase III (1990s--2000s)}: Left 4 Dead AI Director, ALEKS, flow-based systems \cite{booth_ai_2009}\cite{chen_flow_2007}\cite{falmagne_knowledge_2013}&
Low: shallow performance metrics and limited diagnostic reasoning. ALEKS is a partial exception. &
High: real-time difficulty adjustment, PID-style control, and flow optimization. &
Optimizes engagement more than learning. No semantic understanding. Cannot diagnose misconceptions. \\
\addlinespace

\textbf{Phase IV (2000s--2010s)}: e-Adventure, StoryTec, GIFT, EMERGO &
Medium: bounded by authored logic, finite-state structures, scripts, and rule-based assessment\cite{gobel_storytec_2008}\cite{mitrovic2009aspire}\cite{moreno_ger_educational_2008} &
Low: pre-authored branching and limited runtime adaptation. &
Authoring ceiling. No learning over time. Feedback remains generic. \\
\addlinespace

\textbf{Phase V (2010s--2020s)}: Khan Academy, Duolingo, learning analytics \cite{ferguson_learning_2012}\cite{settles_trainable_2016}\cite{siemens_penetrating_2011} &
Medium--high: neural knowledge tracing and large-scale data inference. &
Low--medium: offline batch updates and population-level adaptation. &
Timescale mismatch. Cannot adapt in the moment for an individual learner. \\
\addlinespace

\textbf{Phase VI (2020--present)}: Khanmigo, MATHia with RL, Duolingo Max \cite{kasneci_chatgpt_2023}\cite{walkington_implications_2025}\cite{yan_practical_2023}&
High: LLM reasoning, retrieval-augmented generation, and multi-agent coordination. &
High: real-time policy updating and dynamic content generation. &
Explainability issues, computational cost, hallucination risk, and limited validation evidence. \\
\bottomrule
\end{tabularx}
\end{table*}

\subsubsection{Phase I (1969–1980s): Parameter-Control Computer-Assisted Instruction}
The earliest adaptive training systems emerged from computer-assisted instruction (CAI) research. They adapted to learners by adjusting simple numerical parameters: presentation speed, task complexity, and time pressure. These parameters were designed in response to basic performance signals such as binary accuracy (correct or incorrect), response latency, and error counts \cite{alessi_computer-based_1985}\cite{bunderson_work_1981}\cite{kelley_what_1969}. Representative systems in this phase included PLATO, TICCIT, and various drill-and-practice platforms that used closed-form control logic with hand-tuned thresholds \cite{reiser_history_2001}\cite{sozcu_history_2013}. For example, a rule might state: “If accuracy drops below 70\%, reduce presentation speed by 20\%.” These systems exhibited minimal instructional intelligence: they lacked learner models beyond aggregate error counts, performed no diagnostic reasoning to distinguish why a learner was struggling, and inferred no cognitive state beyond immediate correctness \cite{bunderson_work_1981}\cite{reiser_history_2001}.

The primary constraint was theoretical. These systems were built on behaviorist foundations: the assumption that learning occurs through stimulus-response associations (a behavior followed by a reward or consequence) rather than through internal reasoning or understanding. As a result, systems were designed to shape observable behavior without reasoning about the learner's underlying knowledge structures, an approach now recognized as inadequate for complex skill acquisition requiring conceptual understanding, transfer, and metacognitive awareness (the ability to monitor and regulate one's own learning) \cite{bransford_how_1999}\cite{koedinger_instructional_2013}.

\paragraph {Lesson carried forward to the AI era} %
Phase I demonstrated that computers can adapt instruction to individual learners, and that even simple threshold-based rules improved outcomes compared to one-size-fits-all delivery. However, these early systems revealed a critical limitation: adaptation was driven primarily by observable performance metrics rather than by an understanding of the learner's underlying knowledge state. Contemporary AI systems face the same risk. Reinforcement learning (RL) policies (described in detail in Section 3.2) optimize whatever performance signals they are given as targets. If those signals reflect surface behavior rather than genuine learning, the system may improve measurable outcomes without supporting real understanding. The lesson, recognized more than fifty years ago, remains central: the targets that adaptive systems optimize must be aligned with learning, not just performance.

\subsubsection{Phase II (1980s–1990s): Model-Based Intelligent Tutoring Systems}
Intelligent Tutoring Systems (ITS) represented a substantial increase in instructional intelligence through explicit computational modeling of learner knowledge states, misconception patterns, and pedagogical strategies \cite{anderson_cognitive_1995}\cite{vanlehn_behavior_2006}. Representative systems, including Cognitive Tutors, AutoTutor, SHERLOCK, SimStudent, Betty's Brain, and Apprentice Tutor Builder, implemented three key architectural components: (1) a domain model, encoding expert knowledge about the subject as a structured map of concepts and skills; (2) a learner model, tracking each individual's knowledge state using Bayesian Knowledge Tracing (BKT) which is a probabilistic method that estimates the likelihood a learner has mastered each skill based on their history of correct and incorrect responses \cite{corbett_knowledge_1995}; and (3) a pedagogical module, implementing strategies for deciding what to teach next, how to respond to errors, and how to provide hints \cite{aleven_instruction_2017}\cite{graesser_autotutor_2004}\cite{lesgold_sherlock_1988}. Phase II thus achieved high instructional intelligence but low-to-medium adaptivity: systems could diagnose learner states with considerable sophistication, but their ability to respond rapidly, flexibly, and continuously remained limited by the computational cost of probabilistic inference and the effort required to build each system \cite{brusilovsky_user_2007}\cite{vanlehn_relative_2011}.

\paragraph {Lesson carried forward to the AI era.} %
Phase II established that diagnostic depth matters: systems that reason about why learners err, rather than focusing merely on the fact that they err, can produce substantially better outcomes. Large language models inherit this aspiration while dissolving the knowledge engineering bottleneck. LLMs do not require months of expert effort to encode each new domain. However, an LLM-based tutor without reinforcement learning replicates Phase II’s other weakness: strong performance at the individual interaction level but limited systematic control over the learning trajectory. Semantic understanding of learner responses is necessary but not sufficient; it must be coupled with systematic pedagogical control across sessions.

\subsubsection{Phase III (1990s–2000s): Dynamic Difficulty Adjustment}
Emerging from commercial game development rather than educational technology research, Dynamic Difficulty Adjustment (DDA) systems achieved high adaptivity with low instructional intelligence. They implemented continuous real-time control loops with automated feedback mechanisms that measure a performance signal, compare it to a target, and adjust a parameter to reduce the gap. These loops were used to adjust challenge parameters based on player success rates, failure patterns, and engagement signals \cite{chen_flow_2007}\cite{hunicke_ai_2004}. Representative examples, including the Left 4 Dead AI Director (which adjusts enemy density and difficulty based on player stress signals), flow-based adaptation systems, and ALEKS (Assessment and Learning in Knowledge Spaces), demonstrated rapid responsiveness to interaction patterns \cite{falmagne_knowledge_2013}\cite{sweetser_gameflow_2005}.

DDA systems adjusted difficulty, pacing, and content selection within seconds of observing performance changes \cite{andrade_challenge-sensitive_2005}\cite{chanel_emotion_2011}. These adaptations occurred during activity without interrupting flow, defined as the state of engaged, effortful concentration described by Csikszentmihalyi \cite{csikszentmihalyi_flow_1990}. This contrasted with Phase II systems that intervened episodically between problem-solving attempts \cite{shernoff_student_2003}. However, DDA systems exhibited low instructional intelligence: they lacked explicit learner models, performed no diagnostic reasoning about cognitive states, and operated without grounding in learning science principles \cite{plass_foundations_2015}\cite{vanlehn_behavior_2006}. Their optimization targets were engagement, flow, and player satisfaction rather than durable competence development, skill transfer, or conceptual understanding \cite{mayer_computer_2019}\cite{wouters_meta-analysis_2013}. The exception within Phase III was ALEKS, which achieved moderate intelligence through Knowledge Space Theory, a formal mathematical framework that represents prerequisite relationships among knowledge components, indicating which skills must be mastered before others can be attempted \cite{falmagne_knowledge_2013}.
\paragraph{Lesson carried forward to the AI era.} %
Phase III demonstrated that rapid adaptation is technically feasible. However, without grounding adaptation in diagnostic understanding of the learner's cognitive state, engagement optimization can substitute for learning. A system that keeps learners active and challenged is not necessarily a system that teaches them anything durable. High adaptivity without instructional intelligence risks producing systems that are responsive but pedagogically hollow, a risk that applies equally to modern RL-based systems if their reward signals are poorly specified.
\subsubsection{Phase IV (2000s–2010s): Structured Authoring and Modular Platforms}
Phase IV systems addressed developer productivity through structured authoring environments and modular instructional frameworks. Representative platforms, such as e-Adventure, StoryTec, GIFT, Moirai, VBS3, ELEKTRA, EMERGO, and ASPIRE, introduced separation of concerns as a design principle: the simulation engine (which renders the game world), the assessment module (which evaluates learner performance), and the instructional logic (which decides what to do next) were decoupled into independently maintainable components \cite{bellotti_assessment_2013}\cite{lopes_adaptivity_2011}\cite{moreno_ger_educational_2008}\cite{sottilare_design_2013}. Instructional intelligence was bounded by what authors could encode in advance: finite state machines (FSMs, which are rule-based flowcharts where the system transitions between predefined states based on learner actions), structured scenario scripts, rule-based assessment, and discrete difficulty tiers. Adaptivity was low: systems selected among pre-authored content branches rather than generating new content or modifying instructional logic at runtime.

Phase IV systems addressed developer productivity but not the fundamental challenges of scenario exhaustion, knowledge petrification (the inability to update a system's knowledge after deployment), or generic feedback. Learners still memorized finite content libraries; updates required re-authoring; and systems lacked learner models capable of diagnostic reasoning \cite{bellotti_assessment_2013}\cite{torres_examining_2024}. In this sense, Phase IV’s intelligence ceiling was set entirely at design time.
\paragraph{Lesson carried forward to the AI era.} %
Phase IV’s enduring contribution is the principle of separation of concerns, i.e., organizing a system into distinct, independently manageable components rather than building everything into a single monolithic structure. Decoupling simulation, assessment, and instructional logic improved maintainability and enabled independent component validation. The multi-agent architectures discussed in Section 3.3 apply this same principle: Learner Modeling, Pedagogical Policy, Content Generation, and Validation agents are each independently maintainable and auditable. Phase IV established the correct architectural approach; contemporary AI provides the capability to fill it with learned, dynamic intelligence rather than static, authored content.
\subsubsection{Phase V (2010s–2020s): Mastery Learning and Learning Analytics}
Phase V systems introduced substantial intelligence through large-scale telemetry. The systems focused on automatic collection and transmission of data about learner interactions, such as which questions were answered, how long each took, and how often hints were requested through longitudinal data aggregation, and statistical analysis of learner outcomes across entire educational ecosystems \cite{ferguson_learning_2012}\cite{siemens_penetrating_2011}\cite{sozcu_history_2013}. Representative platforms such as Khan Academy, Duolingo with Half-Life Regression spaced repetition (an algorithm that schedules review of material at optimally spaced intervals, with the spacing determined by the learner's forgetting rate for each item), learning analytics dashboards, and mastery-based progression systems demonstrated that computational analysis of interaction data could inform instructional improvement \cite{pardo_using_2017}\cite{settles_trainable_2016}\cite{verbert_learning_2013}. These systems achieved medium-to-high instructional intelligence through neural knowledge tracing models, machine learning models that predict a learner's current skill level from their history of responses. The patterns  were learned from thousands of prior learners rather than hand-coded rules \cite{piech_deep_2015}. Phase V systems also drew on psychometric item response theory (a statistical framework for estimating the difficulty of questions and the ability of learners from their response patterns) \cite{embretson_item_2006}, and machine learning classifiers predicting at-risk learners from historical patterns\cite{baker_educational_2014}\cite{romero_educational_2010}.

Critically, however, this intelligence was longitudinal and population-level: systems learned which problem types predicted mastery and which learners needed additional support, but they did so by analyzing data from prior learners rather than reasoning about the individual currently interacting with the system \cite{sozcu_history_2013}\cite{viberg_current_2018}. Adaptivity remained low-to-medium: intelligence resided in analytics pipelines and periodic curriculum updates rather than in real-time instructional control during active learning sessions \cite{pardo_using_2017}. This created a fundamental mismatch: systems possessed unprecedented data about learning processes but lacked mechanisms to act on that intelligence in real-time \cite{brinton_learning_2014}\cite{sozcu_history_2013}. Learning analytics answered what happened and what predicts success, but not what the system should do right now for this specific learner in this moment.
\paragraph{Lesson carried forward to the AI era.} %
Phase V’s critical legacy is that interaction data is the raw material for learned intelligence. The historical logs of learner responses accumulated by learning analytics platforms are precisely what modern reinforcement learning (RL) algorithms train on. They use these past interaction records to learn which instructional actions tend to produce good outcomes. Every serious game should therefore be instrumented to capture learning analytics, even before AI capabilities are introduced, because that data becomes the foundation for adaptive intelligence as the system architecture matures.
\subsubsection{Phase VI (2020–Present): AI-Enabled Runtime Adaptivity}
Contemporary systems leveraging large language models (LLMs), reinforcement learning (RL), and neural network architectures represent the first generation with the potential to achieve high instructional intelligence and high adaptivity simultaneously within unified architectures operating at interaction timescales \cite{miljanovic_engineering_2023}\cite{mitsea_systematic_2025}\cite{wang_artificial_2023}. Representative implementations include Khanmigo (a tutoring system powered by GPT-4 that engages learners in conversational dialogue), MATHia with RL-based task sequencing, Duolingo Max, Ferguson et al.'s adaptive VR systems, and Battegazzorre et al.’s narrative difficulty adjustment \cite{battegazzorre_enhancing_2025}\cite{ferguson_ai-induced_2022}\cite{kasneci_chatgpt_2023}.

Intelligence is learned rather than engineered: neural knowledge tracing models infer skill mastery from interaction sequences without hand-coded knowledge components \cite{piech_deep_2015}. LLMs provide semantic understanding of unstructured learner responses. They can interpret the intended meaning of a learner's typed or spoken explanation rather than merely compare it to a predefined list of acceptable answers, enabling open-ended dialogue and context-sensitive feedback without pre-authored templates \cite{brown_language_2020}\cite{mitsea_systematic_2025}\cite{openai_gpt-4_2023}. Multi-agent architectures decompose instructional reasoning into specialized software components: learner modeling, pedagogical strategy, content generation, and safety validation. The agents can coordinate through structured protocols rather than brittle rule systems \cite{wu_autogen_2023}. RL policies optimize task sequencing, difficulty calibration, and intervention timing based on observed trajectories \cite{ferguson_ai-induced_2022}\cite{li_deep_2022}.

Phase VI systems nonetheless remain nascent, with substantial limitations that distinguish them from mature, deployment-ready architectures. Three challenges concern the reliability of individual AI components: (1) explainability: neural models produce predictions without showing their reasoning, making it difficult for instructors and developers to understand or verify why a particular instructional decision was made \cite{yan_practical_2023}; (2) hallucination risk: LLMs sometimes generate content that sounds fluent and authoritative but is factually incorrect, without any built-in mechanism to flag this \cite{kasneci_chatgpt_2023}\cite{yan_practical_2023}; and (3) prompt sensitivity: the same learner input, phrased slightly differently, may produce substantially different LLM outputs, creating reliability and reproducibility challenges. Three further challenges concern system-level deployment: (4) computational cost: running LLMs, training RL systems, and maintaining the required infrastructure involves substantial ongoing expense, and economic viability at scale has not yet been demonstrated; (5) validation complexity: because Phase VI systems generate content dynamically and update continuously, traditional software testing is insufficient to guarantee safety or quality \cite{wu_autogen_2023}; and (6) limited large-scale evidence: most Phase VI systems remain research prototypes; rigorous controlled studies measuring skill transfer over time remain scarce \cite{wang_artificial_2023}.
\paragraph{Lesson carried forward to the AI era.} %
Phase VI inherits the accumulated risks of all prior phases simultaneously: Phase I’s reward-hacking problem (RL policies must be optimized for genuine transfer of learning, not surface accuracy); Phase II’s opacity challenge (neural models are harder to interpret than explicit production rules); Phase III’s engagement trap (learning outcomes, not engagement metrics, must be validated); Phase IV’s architectural lesson (modularity enables independent validation of each component); and Phase V’s data dependency (learned policies are only as good as the interaction data they train on). Addressing these inherited risks requires defining the evidence thresholds that prevent AI-era systems from repeating the field's decades-long pattern of producing technically impressive but educationally inconsistent systems.

\subsection{Synthesis: Architectural Implications for AI-Based Serious Games}

Across five decades and six architectural phases, adaptive training systems oscillated between intelligence-heavy designs that sacrificed responsiveness (Phase II) and adaptivity-heavy designs that sacrificed diagnostic depth (Phase III), with intermediate solutions either separating the dimensions temporally (Phase V analytics informing later curriculum design) or limiting both through static authoring (Phase IV). This enduring trade-off reflected fundamental computational constraints that prior technologies could not overcome: methods capable of deep learner modeling required too much processing time to operate in real time, while systems designed for rapid response lacked the computational resources for complex cognitive reasoning. The historical arc reveals an upper limit on achievable intelligence: an architectural constraint not correctable by better implementation alone. This establishes precisely why contemporary AI technologies represent a potential resolution rather than incremental improvement. Crucially, the historical evolution documents a design-constraint inheritance chain: each phase solved one problem and introduced a new one. AI does not escape this pattern; it inherits all unsolved problems simultaneously.

\section{AI Approaches to Integrating Intelligence and Adaptivity}
\label{sec:3}
The preceding section traced the historical arc that explains why coordinating instructional intelligence and adaptivity has proven so difficult and established why contemporary AI technologies represent a potential resolution. This section examines three AI approaches in sequence: large language models (Section 3.1), which extend the instructional intelligence axis by enabling systems to understand and respond to learner language without manual knowledge engineering; reinforcement learning (Section 3.2), which extends the adaptivity axis by enabling systems to learn instructional policies from interaction data; and agent-based architectures (Section 3.3), which provide a coordination mechanism for combining both capabilities within a single system. Each section analyses what the approach contributes and where its critical limitations lie.
\subsection{Large Language Models Extend the Intelligence Axis: Semantic Understanding Without Manual Knowledge Engineering}
Historically, instructional intelligence was bounded by information that could be manually encoded through knowledge engineering, a labor-intensive process. Human domain experts worked with software developers to formalize a body of knowledge into structured representations that a computer can reason about. This involved encoding domain knowledge as concept hierarchies and prerequisite maps, representing learner models as skill mastery probabilities and catalogues of known misconceptions, and specifying pedagogical strategies as production rules: explicit "if this situation, then do this" instructions \cite{anderson_cognitive_1995}\cite{anderson_taxonomy_2001}\cite{vanlehn_behavior_2006}. Each domain required sustained expert effort; systems remained brittle when learners deviated from anticipated paths, and scaling to new content areas demanded proportional expert investment \cite{aleven_instruction_2017}\cite{self_defining_1998}. The fundamental constraint was that every concept, skill, error pattern, and instructional strategy that the system could reason about had to be explicitly enumerated and encoded prior to deployment.

Large language models (LLMs) address this constraint through learned representations. They are AI systems trained on large collections of text to develop a broad understanding of language, concepts, and their relationships. Rather than being programmed with explicit rules, LLMs develop internal mathematical structures that capture how words, concepts, and contexts relate to one another. They acquire these structures by processing billions of examples of human-written text \cite{brown_language_2020}\cite{kasneci_chatgpt_2023}\cite{openai_gpt-4_2023}. For serious games, this enables several capabilities previously requiring manual authoring:
\begin{enumerate}
\item{Natural language interaction:}
LLMs can interpret unstructured learner input (free-text explanations, spoken dialogue, partial solutions) and thereby reveal reasoning processes without requiring pre-specification of acceptable response formats or anticipated phrasing variations  \cite{brown_language_2020}\cite{kasneci_chatgpt_2023}. In healthcare simulation, this capability allows virtual patients to respond naturally to open-ended history-taking questions rather than requiring learners to select from multiple-choice menus \cite{cook_virtual_2025}.

\item{Context-sensitive explanation generation:}
Rather than selecting from pre-authored feedback templates, LLMs can generate explanations tailored to the specific error a learner just made, the context of the scenario, and what the system knows about that learner's prior performance \cite{nye_generative_2023}. In procedural skill training, this means feedback can address not just what went wrong but why that approach is problematic, how it relates to underlying principles, and what alternative strategies would be more appropriate \cite{hattie_power_2007}\cite{shute_focus_2008}.

\item{Grounded content generation:}
retrieval-augmented generation (RAG) is a technique in which an LLM's response is informed by retrieving relevant passages from a verified knowledge base before generating text. Through RAG, LLMs can combine their language capabilities with access to validated domain sources such as medical protocols, technical documentation, and assessment rubrics, enabling scenario generation that maintains factual accuracy while introducing surface variation  \cite{kasneci_chatgpt_2023}\cite{lewis_retrieval-augmented_2020}.

\item{Metacognitive scaffolding:}
LLMs can generate Socratic prompts (questions designed to guide the learner to discover an answer rather than simply providing it), reflective questions, and worked-example variations that support metacognition, defined as the learner's ability to monitor their own understanding, notice gaps, and regulate their learning strategies \cite{bransford_how_1999}\cite{kasneci_chatgpt_2023}\cite{nye_generative_2023}\cite{shute_focus_2008}.
\end{enumerate}
\subsubsection{Critical Limitations: LLMs Provide Reactive Intelligence Without Proactive Pedagogical Control}
While LLMs substantially expand what a system can understand and generate, they do not inherently address the adaptivity problem. A standalone LLM-based tutor primarily operates at the level of the current interaction: it can interpret what the learner just said and generate a contextually appropriate response, but it does not maintain a persistent model of the learner's evolving knowledge state. Such a model requires a running record of what has been mastered, what misconceptions remain, and how understanding has changed across sessions \cite{nye_generative_2023}\cite{wang_artificial_2023}. Without additional components, such systems cannot track which concepts a learner has repeatedly struggled with, reliably distinguish genuine understanding from memorized responses, or adjust the overall instructional trajectory based on longitudinal patterns of performance. In addition, LLM outputs are sensitive to prompt formulation, i.e., the exact wording and structure of the text passed to the model \cite{brown_language_2020}. Behaviorally similar learner inputs can produce substantially different outputs depending on how the system frames the question. This creates reproducibility and reliability challenges that are not present in deterministic rule-based systems. The discrepancy is particularly concerning in high-stakes training domains where consistency of instructional decision-making matters \cite{kasneci_chatgpt_2023}.

In terms of the historical arc, standalone LLM tutors replicate Phase II’s profile: strong instructional intelligence at the single-interaction level, but without systematic control over the learning trajectory over time. Early LLM-based educational systems accordingly showed mixed results: impressive conversational capabilities coexisted with concerns about pedagogical aimlessness, the tendency for conversation to drift without a coherent instructional objective \cite{kasneci_chatgpt_2023}\cite{wang_artificial_2023}. An additional safety concern is that LLMs may generate confident, fluent content that is factually incorrect or pedagogically inappropriate. This is a risk that is particularly serious in high-stakes training domains such as healthcare or defense \cite{afroogh_trust_2024}\cite{yan_practical_2023}.
\subsection{Reinforcement Learning Extends the Adaptivity Axis: Learned Policies for Instructional Control}
Early adaptive systems relied on hand-authored control logic: explicit IF–THEN rules triggering parameter adjustments when performance crossed pre-defined thresholds (Phase I), or engagement heuristics implementing DDA based on flow theory principles (Phase III). These approaches limited adaptivity to patterns anticipated by designers. Reinforcement learning (RL) replaces hand-authored control logic with policies, defined as decision-making strategies, that are learned from experience rather than explicitly programmed \cite{sutton_reinforcement_2018}. In the RL paradigm, a software agent learns by trial and error: it observes the current state of the learning environment (which might include the learner's recent performance, their inferred knowledge state, and the current task), selects an instructional action, receives a reward signal indicating how good that action was, and gradually develops strategies that maximize cumulative reward over time \cite{silver_mastering_2017}\cite{vinyals_grandmaster_2019}. The analogy is an experienced teacher who has learned through years of practice which instructional moves tend to work well, rather than following a fixed script.

For serious games, RL reframes instructional adaptation as a sequential decision-making problem: given current beliefs about learner state, what instructional action should the system take to optimize learning objectives \cite{ferguson_learning_2012}\cite{li_deep_2022}\cite{narvekar_curriculum_2020}? Actions include task sequencing (determining which scenario the learner should attempt next), difficulty calibration (adjusting task parameters to current performance), intervention timing (deciding when to provide hints versus allowing productive struggle, the educationally beneficial experience of working through a challenge without immediate assistance), and scaffolding fading (gradually withdrawing support as learners gain competence, allowing them to develop independence) \cite{holstein_student_2018}\cite{koedinger_exploring_2007}\cite{narvekar_curriculum_2020}\cite{wood_role_1976}. Crucially, RL policies operate at interaction timescales, enabling genuinely real-time adaptivity \cite{li_deep_2022}.
\subsubsection{Critical Limitations: RL Optimizes Observables, Not Cognitive States}
RL addresses the control problem of how to make instructional decisions in real time. It does not inherently address the epistemic problem: what does the learner know, misunderstand, or need to develop conceptually \cite{koedinger_instructional_2013}\cite{plass_foundations_2015}? The fundamental constraint is that RL agents optimize whatever objectives are specified through reward functions (the mathematical specification of what counts as a good outcome \cite{narvekar_curriculum_2020}), and whatever information about the learner is captured in the system's state representation. If these are defined poorly, RL produces adaptive behavior that appears intelligent while being pedagogically hollow.

Three failure scenarios illustrate this risk. First, reward hacking through memorization: an RL policy may learn to present repeated similar scenarios because learners achieve high accuracy through pattern matching. The system is "rewarded" for high accuracy, so it learns to produce easily achievable accuracy without distinguishing genuine skill mastery from superficial memorization \cite{koedinger_instructional_2013}\cite{li_deep_2022}. Second, engagement optimization over learning: if the reward signal incorporates completion rates or time-on-task, the policy may learn to present easier, more engaging content that keeps learners active rather than challenging content that produces learning gains \cite{hunicke_ai_2004}\cite{wouters_meta-analysis_2013}. Third, scaffolding dependency: a policy that provides extensive hints to produce high immediate success rates may create dependency where learners perform well with support but fail without it \cite{aleven_instruction_2017}\cite{kapur_productive_2023}.

More fundamentally, defining what "success" means in a way that captures genuine learning remains an open research challenge. There is no agreed methodology for specifying pedagogically valid reward functions for complex skill domains, and proxies such as assessment scores, task completion, or time-on-task each introduce their own biases. The solution is not to abandon RL but to recognize its architectural role: RL provides adaptive control mechanisms, but it requires rich information about the learner's cognitive state and carefully designed reward functions to optimize pedagogically meaningful objectives \cite{narvekar_curriculum_2020}. RL answers how to adapt but not what to adapt to. This creates an architectural dependency: effective RL-based instructional adaptation requires external systems that provide cognitive state inference, which in turn requires more advanced AI models.
\subsection{Agentic Architectures: Coordinating Intelligence and Adaptivity}
The preceding sections establish that LLMs and RL each extend one dimension of the challenge independently. LLMs provide rich instructional intelligence at the level of the current interaction but do not inherently maintain persistent learner models or pursue long-term pedagogical objectives. RL provides adaptive control over instructional decisions but optimizes whatever performance signals it is given, without an inherent understanding of what the learner is thinking or why they are struggling. Neither technology, used alone, resolves the historical trade-off between intelligence and adaptivity identified in Section 2. What is needed is a mechanism to coordinate these capabilities at runtime so that each technology contributes what it does well while compensating for the limitations of the other.

Agent-based architectures provide one such coordination mechanism. In computing, a software agent is a program that perceives inputs from its environment, maintains an internal state representing its current beliefs or knowledge, makes decisions autonomously in pursuit of a specified objective, and produces outputs, such as data, decisions, or generated content, that affect its environment or are passed to other components \cite{russell_artificial_2020}\cite{wooldridge_introduction_2009}. Unlike a simple function that maps inputs to outputs and then terminates, an agent is persistent: it runs continuously, updates its state based on incoming information, and acts over time rather than at a single moment. A multi-agent system is a software architecture in which two or more such agents operate concurrently, each specializing in a distinct function, and coordinate by passing structured information to one another \cite{wooldridge_introduction_2009}. Building a single monolithic system that attempts to perform all instructional functions simultaneously has proven technically infeasible, as the historical arc demonstrates. Agent-based approaches instead distribute responsibility across components that can be developed, tested, validated, and updated independently \cite{wooldridge_intelligent_1995}\cite{wu_autogen_2023}.

Applied to serious games, agent-based thinking offers a way to decompose the instructional system into roles that correspond to recognizable functions in human-mediated training: a component that tracks and diagnoses what the learner knows (analogous to a clinical supervisor assessing a trainee's competence), a component that decides what instructional action to take next (analogous to a curriculum advisor), a component that generates the content the learner will experience (analogous to a scenario author), and a component that checks the quality and safety of that content before it reaches the learner (analogous to a subject matter reviewer) \cite{vanlehn_relative_2011}\cite{woolf_building_2008}\cite{wooldridge_introduction_2009}. These functions are not new. They have been present, in some form, in all phases of the historical arc, but prior systems required them to be encoded manually at design time \cite{lopes_adaptivity_2011}\cite{sottilare_design_2013}. Agent-based AI architectures make it possible, in principle, to instantiate these functions as dynamic, data-driven components that operate continuously during a live learning session.

The key capability that agent-based architectures offer for serious games is runtime coordination of intelligence and adaptivity. Where Phase II ITS systems achieved deep instructional intelligence but adapted slowly, and Phase III DDA systems adapted rapidly but without diagnostic depth, an agent-based system can in principle achieve both simultaneously: one agent continuously updates a rich model of the learner's knowledge state, another uses that model as input to make rapid instructional decisions, and a third generates content appropriate to those decisions, all within timescales compatible with interactive gameplay. The instructional interaction loops, which perceive the learner's action, update the learner model, decide on an instructional response, generate content, validate it, and deliver it, can in principle complete within the natural pauses and transitions of a game session rather than requiring the kind of offline analysis or curriculum redesign that characterized earlier phases \cite{ferguson_ai-induced_2022}\cite{holstein_student_2018}\cite{wu_autogen_2023}.

Agent-based approaches also offer a degree of modularity that has practical advantages for serious game development and maintenance. Because each agent has a well-defined interface specifying what information it receives and what it produces, individual components can be updated or replaced without rebuilding the entire system. A domain expert can update the knowledge base that the content generation component draws on without touching the learner modeling component. An instructor can configure the objectives that the instructional policy component optimizes without modifying the content generation component. When the system behaves inappropriately, developers can in principle identify which component produced the problematic output rather than debugging an opaque, all-in-one system. This modularity is consistent with the principle of separation of concerns that Phase IV authoring platforms introduced and that the historical arc identifies as architecturally important. The principle can now be extended to intelligent, learned components rather than static authored rules.

The range of agent-based configurations that have been proposed or implemented varies considerably in complexity and scope. At the simpler end, a two-agent design might pair a learner modeling component that tracks skill mastery with a policy component that selects the next scenario, extending Phase V analytics with real-time instructional control. At the more complex end, research systems have explored architectures in which multiple specialized agents coordinate across scenario generation, affective state monitoring, dialogue management, and safety validation simultaneously. No single configuration has been established as standard; the appropriate architecture depends on the training domain, the nature of the skills being developed, the available data infrastructure, and the computational resources the deployment context supports. This flexibility is itself a feature: agent-based thinking provides a vocabulary for articulating the functional requirements of an intelligent adaptive serious game without prescribing a fixed technical solution  \cite{wooldridge_introduction_2009}\cite{wu_autogen_2023}.

\subsubsection{Critical Limitations: Coordination Introduces Complexity, Opacity, and Unresolved Validation Challenges}
Agent-based architectures address the coordination problem, but they introduce a new set of challenges that are distinct from those of LLMs and RL individually and that have implications for serious game deployment in high-stakes training contexts.

The first challenge is coordination overhead. Each agent must receive inputs from other agents, process them, and produce outputs in time for the next agent in the pipeline to act. The total latency of this chain determines how quickly the system can respond to a learner's action. LLM-based content generation is the primary bottleneck: producing even a short feedback message typically requires 500 milliseconds to two seconds depending on context length and model size \cite{kasneci_chatgpt_2023}. Across a multi-agent pipeline, total response times may exceed what is compatible with the moment-to-moment interaction of an immersive game session. Practical implementations therefore require asynchronous pipelines, meaning that content generation occurs in the background while the learner continues interacting with the game, with instructional responses delivered at natural transition points rather than immediately after each learner action. Designing these pipelines requires careful attention to when and how the game pauses or transitions to accommodate instructional delivery.

The second challenge is emergent behavior that can be traced to system-level outcomes that arise from the interaction of multiple agents but were not explicitly programmed into any individual component \cite{wu_autogen_2023}. When a learner modeling agent, a policy agent, and a content generation agent each make decisions based on their own inputs and objectives, the combined behavior of the system can produce outcomes that none of the individual components would have produced alone, and that system designers did not anticipate. This is distinct from the kind of bugs or failures that arise in deterministic systems (where the same input always produces the same output and errors can be traced to specific lines of code), because the behavior of each agent is learned and probabilistic, and the interaction between agents is dynamic. Emergent behaviors may be beneficial (the system discovers effective instructional strategies that were not explicitly programmed) or harmful (the system develops patterns of behavior that are superficially coherent but pedagogically inappropriate). Detecting and correcting them requires evaluation methodologies that go beyond traditional software testing  \cite{wu_autogen_2023}.

The third challenge is transparency and instructor trust. A serious game instructor overseeing a training session needs to understand, at a meaningful level, what the system is doing and why. In a traditional serious game, this is straightforward: the instructor knows what scenarios exist, what the branching logic is, and what feedback messages are pre-written. In an agent-based system with LLM-generated content and RL-derived policies, the instructional decisions are the product of learned models whose internal reasoning is not directly interpretable. An instructor who sees a learner receive unexpected feedback, or who observes the system making an instructional choice that seems pedagogically inappropriate, may have no practical means to understand why that decision was made or to prevent it from recurring. This opacity is not merely an interface problem. Rather, it reflects a fundamental property of the underlying AI components. Addressing it requires deliberate architectural investment in logging, explanation, and instructor control mechanisms that are not naturally provided by the AI components themselves \cite{miljanovic_engineering_2023}\cite{yan_practical_2023}.

The fourth challenge is validation complexity. Validating a serious game traditionally means reviewing its content which involves checking that scenarios are accurate, feedback is appropriate, and assessment criteria are correct before deployment. Agent-based systems with generative content cannot be validated in this way, because there is no fixed content to review: the system produces content dynamically in response to each individual learner. Validation must instead assess the process by which the system makes decisions: whether the learner model accurately reflects learner knowledge across a representative range of learner types and behaviors, whether the policy agent makes pedagogically sound decisions across a representative range of learner states, and whether the content generation component produces accurate and appropriate content across a representative range of instructional contexts. This requires simulation-based evaluation (running the system against synthetic learner profiles), expert monitoring during live deployments, and ongoing auditing throughout the system's operational life. The validation burden for agent-based serious games is substantially higher than for traditional systems and does not end at deployment \cite{wu_autogen_2023}\cite{yan_practical_2023}.

Finally, it is important to situate agent-based architectures accurately within the current state of the field. While conceptually coherent and technically grounded, these architectures remain largely aspirational in fully integrated form. Existing systems implement individual components such as LLM-based dialogue, RL-based task sequencing, and neural knowledge tracing, but no deployed serious game has yet demonstrated all of these components operating simultaneously at interaction timescales in a validated high-stakes training context. Agent-based thinking currently provides more of a productive design framework and research direction than a proven deployment model. The translation from concept to practice requires sustained investment in both research and engineering infrastructure that the field has not yet completed.

Agent-based architectures represent a principled solution to the difficulty of combining instructional intelligence and real-time adaptivity within a single coherent system. By distributing responsibility across specialized components that communicate through explicit interfaces, these architectures extend the principle of separation of concerns that Phase IV established and apply it to dynamic, learned instructional functions rather than static authored content. However, introducing agent-based coordination does not resolve the inherited risks from prior phases; it inherits them all simultaneously, while adding new challenges of emergent behavior, transparency, and validation complexity. Whether the benefits of coordination outweigh these challenges in any specific serious game deployment depends on the training domain, the available data infrastructure, the regulatory requirements governing the training context, and the investment the development team can sustain in ongoing monitoring and maintenance. These are ultimately empirical questions, and the empirical evidence required to answer them rigorously remains to be developed.

\section{Conclusion}
\label{sec:4}

This chapter examined how instructional intelligence and adaptivity, two system properties that have historically proven difficult to realize simultaneously, may be better integrated through contemporary AI architectures. The historical review presented in Section 2 reveals a recurring pattern across more than five decades: systems tended to emphasize either diagnostic depth or runtime responsiveness, but rarely both. Intelligent Tutoring Systems achieved sophisticated learner modeling but were constrained by knowledge engineering bottlenecks and limited real-time responsiveness. DDA systems achieved rapid runtime adaptation but relied on shallow performance indicators, optimizing engagement rather than durable learning outcomes. Intermediate architectures, such as structured authoring platforms and learning analytics systems, addressed specific constraints while introducing new ones.

The analysis suggests that these trade-offs reflected architectural constraints within earlier generations of training systems rather than merely suboptimal design choices. The computational methods required for deep learner modeling were incompatible with real-time interaction constraints. Systems designed for rapid response lacked the resources for complex cognitive reasoning. Contemporary AI technologies change some of these constraints but do not eliminate them.
LLMs expand representational capacity by enabling systems to interpret natural language input, generate contextually sensitive explanations, and produce adaptive feedback without relying on manually authored knowledge structures. However, LLMs primarily operate at the level of the current interaction. Without additional architectural components, they do not maintain persistent learner models, track learning trajectories over time, or implement systematic pedagogical strategies. RL provides a complementary capability by enabling systems to learn instructional policies (decision-making strategies for choosing tasks, adjusting difficulty, and timing support) that improve through experience. Yet RL systems optimize the reward signals they are given: if those signals reflect short-term performance or engagement metrics, the resulting policies may produce adaptive behavior that appears effective while failing to support deeper conceptual learning.

This observation reinforces a central design principle: instructional intelligence and adaptivity must be integrated within architectures that align learner modeling, instructional policy, and content generation with meaningful learning objectives. Multi-agent architectures provide one promising approach by decomposing instructional systems into specialized components that are independently maintainable and auditable, preserving modularity and transparency. Responsible integration additionally requires mechanisms for instructor oversight, transparent decision-making, learner data governance, and the capacity for human override — criteria that current systems have not yet systematically addressed.

Despite the technical advances, important limitations and challenges remain. AI-enabled serious games introduce risks related to system validity, explainability, economic cost, and governance. Neural network (deep learning) models often produce predictions without interpretable reasoning. Generative models introduce the possibility of inaccurate or misleading outputs. RL policies can produce unintended strategies if reward functions are poorly specified. Large-scale empirical evidence demonstrating long-term learning gains and transfer from AI-enabled serious games remains limited. These challenges highlight the need for rigorous validation, transparent system design, and evaluation frameworks that extend beyond engagement metrics to measure learning effectiveness and skill transfer.

Taken together, the historical arc presented in this chapter suggests that AI integration represents a meaningful shift in how instructional intelligence can be embedded within serious games. Earlier phases progressively addressed specific limitations through parameter-based adaptation, cognitive diagnosis, engagement optimization, modular authoring, and large-scale learning analytics. Contemporary AI systems inherit insights and constraints from each of these phases. Their success will therefore depend not only on technical capability but on whether they address the accumulated lessons of prior generations of adaptive training systems.
Several research priorities emerge from this analysis. First, rigorous empirical evidence is needed regarding the educational impact of AI-enabled serious games, particularly longitudinal studies measuring skill transfer rather than engagement or short-term performance. Second, transparent and auditable architectures are needed so that instructors and developers can inspect, understand, and appropriately trust system behavior. Third, clear design principles are needed to align AI capabilities with established theories of learning and training, including explicit criteria for specifying reward functions in RL-based systems and for evaluating the pedagogical appropriateness of LLM-generated content. If implemented responsibly and evaluated rigorously, AI-enabled architectures have the potential to move serious games closer to a longstanding goal: systems capable of providing individualized instruction, adaptive challenge, and meaningful feedback at scale, while maintaining pedagogical integrity and learner trust.

\printbibliography
\end{document}